# An Improved Expectation Maximization Algorithm

Chen Fuqiang

Department of computer science and technology

College of Electronics & Information Engineering

Tongji University

Shanghai, China, 201804

2012fuqiangchen@tongji.edu.cn

Abstract

In this paper, we firstly give a brief introduction of expectation maximization (EM) algorithm, and then discuss the initial value sensitivity of expectation maximization algorithm. Subsequently, we give a short proof of EM's convergence. Then, we implement experiments with the expectation maximization algorithm (We implement all the experiments on Gaussion mixture model (GMM) ). Our experiment with expectation maximization is performed in the following three cases: initialize randomly; initialize with result of K-means; initialize with result of K-medoids. The experiment result shows that expectation maximization algorithm depend on its initial state or parameters. And we found that EM initialized with K-medoids performed better than both the one initialized with K-means and the one initialized randomly.

1 Introduction

Maximum likelihood estimation (MLE) is a parameter estimation method widely applied in statistics [1]. In maximum likelihood estimation, the condition is that we are given a set of independently identically distributed (i.i.d.) points or samples, and we know the specific form of the distribution where the samples are drawn from beforehand. Maximum likelihood estimation finds the real parameters in the distribution by maximizing the probability, which is transformed to a logarithmic form, and thus it's much easier to compute the derivative of the object function with respect to (w.r.t.) the unknown parameters.

However, in most cases, we don't know the label of each sample we get. For example, if we are given two coins, A and B, and then we randomly choose one of them to flip [2]. The upward side of the coin may be head or tail. Suppose that we don't know which coin is chosen in each flipping, and our object is to estimate the probability of head upward for each coin under the condition that for each flipping we don't know which coin is chosen. This can be solved by expectation maximization algorithm, which mainly deals with the parameter estimation with latent or hidden variables. And in the above coin flipping problem, the hidden variable is which coin we choose in each flipping, or rather, we don't know the probability of choosing one specific coin, A and B.

EM algorithm has been applied in various areas. Shepp, L. A. et al. (1982) [3] resorted to EM algorithm indirectly to emission tomography Image Reconstruction. Feder, M. et al. (1989) [4] applied EM algorithm for maximum likelihood Active Noise Cancellation (ANC). Carson, C. et al. (2002) [5] used EM algorithm for image segmentation. Kriegel, H. P. et al. (2006) [6] found that for multi-instance problem, EM algorithm performed better than k-medoid algorithm on three real world data sets. To our knowledge, there are no researchers who initialize EM algorithm with k-medoids. And our experiment result show that EM algorithm initialized with k-medoids performs better than that initialized with k-means or initialized randomly.

This paper is organized as follows. In section 2, we introduce EM briefly. In section 3, we analyze the sensitivity of EM algorithm on the initialization. And in section 4, we present the convergence of EM algorithm. In section 5, we show the experiment result and gives an analysis. Finally, in section 6, we conclude our work and point out the future direction of our work.

2 Brief introduction of EM

In this section, we introduce the expectation maximization algorithm mathematically and generally.

Suppose that we are given n d-dimensional samples $X=\{x_1, ..., x_n\}$ collected from K clusters, groups or classes, where $x_i \in R^d$. And for $i=1, ..., n$, we don't know which cluster $x_i$ comes from, i.e., the labels of all the samples are latent variables. Our object is to estimate the parameter θ in the probability density function (pdf) $p(X; θ)$ which maximizes the probability density function. Here the θ usually represents a parameter set. For example, for one dimensional normal distribution, θ = {$\mu_j$ , $\sigma_j$ ; j = 1 , ... , K}, where $\mu_j$ represents the mean of all the samples from the j-th cluster, and $\sigma_j$ represents the variance of all the samples from the j-th cluster.

Generally, for convenience, and also because the function *ln*(x) increases monotonously, which makes the equivalence of maximizing the two function, $L(\theta) = p(X; \theta)$ and $l(\theta) = \ln p(X; \theta)$, we chose to maximize the latter one instead.

In this paper, we discuss expectation maximization algorithm for clustering or classification, and we use y to represent the latent variable. Generally, y can take on some numbers making up a set *Y* composed of several integers, such as *Y*={1, 2, ... , K}. Then, by the complete probability formula, we can get

$$p(X; \theta) = \sum_{j=1}^{k} p(X; y_j, \theta) p(y_j; \theta),$$

where $p(y_j; \theta)$ denotes the sum of all the probability of sample $x_j$ belongs to the j-th cluster. So

$$l(\theta) = \ln \sum_{j=1}^{k} p(X; y_j, \theta) p(y_j, \theta),$$

and our object is to maximize $l(\theta)$ . It's obvious that $l(\theta)$ is a logarithmic function of a summation of k function, and usually it's hard to compute the maximum of $l(\theta)$ directly. Then we can chose an initial value $\theta_l$ for θ, and then

$l(\theta) - l(\theta_l)$

$= \ln p(X; \theta) - \ln p(X; \theta_l)$

$= \ln [p(X; \theta) / p(X; \theta_l)]$

$= \ln \{[\sum_{j=1}^{k} p(X; y_j, \theta) p(y_j; \theta)] / p(X; \theta_l)\}$

$= \ln \sum_{j=1}^{k} [p(X; y_j, \theta) p(y_j; \theta) p(y_j; X, \theta_l) / p(X; \theta_l) p(y_j; X, \theta_l)]$

$\geq \sum_{j=1}^{k} \{p(y_j; X, \theta_l) \ln [p(X; y_j, \theta) p(y_j; \theta) / p(X; \theta_l) p(y_j; X, \theta_l)]\}.$

We deduce the last step (inequality) by the famous Jenson's inequality, and

$$\sum_{j=1}^{k} p(y_j; X, \theta_l) = 1.$$

Then we can maximize

$$\sum_{j=1}^{k} p(y_j; X, \theta_l) \ln [p(x; y_j, \theta) p(y_j; \theta) / p(X; \theta_l) p(y_j; X, \theta_l)]$$

with respect to θ to find a relative better lower bound than generalized expectation maximization (GEM) algorithm [7], which chose any θ to make $l(\theta)$ increase in each iteration. It's easy to see that in the above formula, the denominator has nothing to do with θ, thus we can neglect it when we maximize the above formula.

And finally we can maximize the following

$$\sum_{j=1}^{k} p(y_j; X, \theta_l) \ln p(X; y_j, \theta) p(y_j; \theta),$$

which in fact is

$$E_{\{y|X, \theta_l\}} [\ln p(X, y; \theta)].$$

As yet we have got the E-step of EM algorithm, which is to compute the above expectation.

The next step in EM algorithm, i.e., M-step, is to maximize the expectation which we get in the E-step with respect to (w.r.t.) θ. Formally, our object is to get the $\theta_{l+1}$:

$$\theta_{l+1} = \mathrm{argmax}_\theta E_{\{y|X, \theta_l\}} [\ln p(X, y; \theta)].$$

We have given an introduction to expectation maximization algorithm in the above.

3 Sensitivity analysis

In this section, we analyze the sensitivity of expectation maximization algorithm with respect to the parameters in the probability density function, with the number of total clusters clapped, i.e., we make the number of clusters an invariant value. Intuitively, if we initialize the parameters with different values in each experiment, the performance of expectation maximization algorithm differ from each other. Here, we analyze the initial value sensitivity or performance according to the experiment, for that it's hard to analyze mathematically directly. Intuitively, if the clusters we deal with satisfy the following conditions:

(1) there are many clusters;

(2) the mean values of any two clusters/groups are close to each other measured by a specific distance metric ( in our paper we consider the Euclidean distance );

(3) the covariance of any cluster is so large that some points/samples of this cluster may be in the cloud formed by another cluster;

For clarity, the readers are referred to Section 4 (i.e. Experiment). Here we only give the analysis intuitively, if you are interested in the proof mathematically, you can develop the proof by yourself.

In conclusion, the function $l(\theta)$ may have some local maximums besides the global maximum, thence if we initialize the parameters with different values, expectation maximization algorithm may converge to a local maximum, unless the initial value are initialized close to the true parameters, which is generally very hard in practice.

4 Convergence analysis

We talk about the convergence of expectation maximization algorithm in this section. It's obvious that the following holds:

$l(\theta_{l+1})$

$\geq l(\theta_l) + \sum_{j=1}^{k} p(y_j; X, \theta_l) \ln [p(X; y_j, \theta_{l+1}) p(y_j, \theta_{l+1}) / p(X; \theta_l) p(y_j; X, \theta_l)]$

$\geq l(\theta_l) + \sum_{j=1}^{k} p(y_j; X, \theta_l) \ln p(X; y_j, \theta_l) p(y_j, \theta_j) p(X; \theta_j) p(y_j; X, \theta_l)\}$

$= l(\theta_l),$

through which we can see that $l(\theta_l)$ is monotonously increasing, and it's obvious that $l(\theta)$ is bounded. So the convergence of expectation maximization algorithm holds by the theorem bounded sequence will be converged in Mathematical Analysis [8].

## 5 Experiment

Before presenting the experiment result, we first give a brief introduction to K-means algorithm which is usually used in clustering [9]. Given n samples in K clusters, the K-means algorithm is an algorithm which give K means for K clusters initially by a specific rule and change the K means by some rule to make the means close to the true means. And in our experiment, we stop the K-means algorithm if all the K means don't change any more. Besides the K-means algorithm, there is another algorithm, K-medoids, which slightly like K-means algorithm. And we also implement experiment with EM algorithm initialized by K-medoids [10].

In our experiment, we mainly deal with the following cases:

(1) four clusters in 2-dimension space;

(2) four clusters in 3-dimension space;

We implement many experiments with K-means algorithm and EM algorithm respectively on Gaussion mixture models, and for the details of Gaussion mixture models the readers are referred to [11].

Here we only give a brief introduction of how to get the estimate of $\mu$ and $\Sigma$. To this end, we can regard $\Sigma^{-1}$ as a generalized reciprocal for matrix of $\Sigma$ and regard $|\Sigma|$ as the generalized absolute value for matrix of $\Sigma$, both regarding $\Sigma$ as a generic variable such as $x$ when we calculation partial derivative with respect to $\Sigma$. For the proof in detail the readers are referred to [11]. In our experiment, we stop when the parameters change smaller than a specific given threshold. The experiment results in the appendix give an intuitive proof of what we analyze in section 2.

Considering that in expectation maximization algorithm if we initialize the parameters with random values, the performance of expectation maximization algorithm is inclined to perform poorly. So we initialize the parameters with the result of K-means algorithms and K-medoids algorithm. We implement experiment 50 times in each case ( 2d space EM not initialized by K-means; 2d space EM initialized by K-means; 3d space EM not initialized by K-means; 3d space EM initialized by K-means; 2d space EM initialized by K-medoids; 3d space EM not initialized by K-medoids ).

The experiment result is given in the appendix. From the six tables, we can conclude that expectation maximization algorithm performs well after combining with K-means algorithm and K-medoids algorithm, i.e., if we initialize the parameters for expectation maximization by the result of K-means and K-medoids algorithm respectively, EM performs better. We find that EM initialized with K-means or K-medoids improve the EM's performance, and in lower dimensions more obvious.

6 Conclusion

In this paper, we first give a brief introduction of expectation maximization algorithm, which is a method different from maximum likelihood estimation, dealing with problems with latent or hidden variables. Then we prove the convergence of the EM algorithm briefly.

On the bases above, we implement experiment with K-means algorithms and expectation maximization algorithm and give some analysis. The experiment result show that both K-means algorithm and expectation maximization algorithm are sensitive to the initial value, so in future some adaptive or generalized algorithm need to be proposed for improve the classification rate or performance. And, in our experiment, we implement experiment of initializing the parameters according to the result of K-means algorithm, the results show that EM performs better than randomly initializing the parameters. Since K-means and K-medoids can improve EM's performance, it's easy to think about how about applying other clustering algorithm to initialize EM algorithm.

Besides, in this paper, we performed EM on artificial data, in future, we will perform EM on real world dataset to test the validity of our algorithm,

Acknowledgment

The author thank Yilu Zhao for helpful discussion.

Appendix

Table 1: Result of EM not initialized by K-means: 2d space

| none well | K-means well /$N_K$ | EM well /$N_E$ | $N_K/N_E$ |
|---|---|---|---|

| 11 | 31 | 27 | 1.15 |

Table 2: Result of EM not initialized by K-means: 2d space

| none well | K-means well /$N_K$ | EM well /$N_E$ | $N_K/N_E$ |
|---|---|---|---|
| 16 | 27 | 33 | 0.82 |

Table 3: Result of EM not initialized by K-means: 3d space

| none well | K-means well /$N_K$ | EM well /$N_E$ | $N_K/N_E$ |
|---|---|---|---|
| 6 | 36 | 33 | 1.09 |

Table 4: Result of EM not initialized by K-means: 3d space

| none well | K-means well /$N_K$ | EM well /$N_E$ | $N_K/N_E$ |
|---|---|---|---|
| 9 | 41 | 41 | 1 |

Table 5: Result of EM not initialized by K-means: 3d space

| none well | K-means well /$N_K$ | EM well /$N_E$ | $N_K/N_E$ |
|---|---|---|---|
| 6 | 36 | 28 | 1.28 |

Table 6: Result of EM not initialized by K-means: 2d space

| none well | K-means well /$N_K$ | M well /$N_E$ | $N_K/N_E$ |
|---|---|---|---|
| 12 | 27 | 31 | 0.87 |

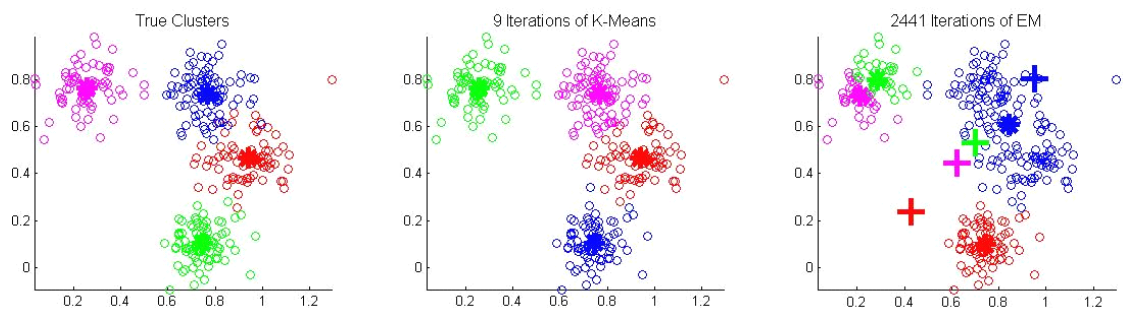

Figure 1: Case of four clusters in 2-dimension space and the *'s represent the mean for every cluster, and the +'s represent the initial mean for EM. In this case, the K-means performs well, while EM performs poorly.

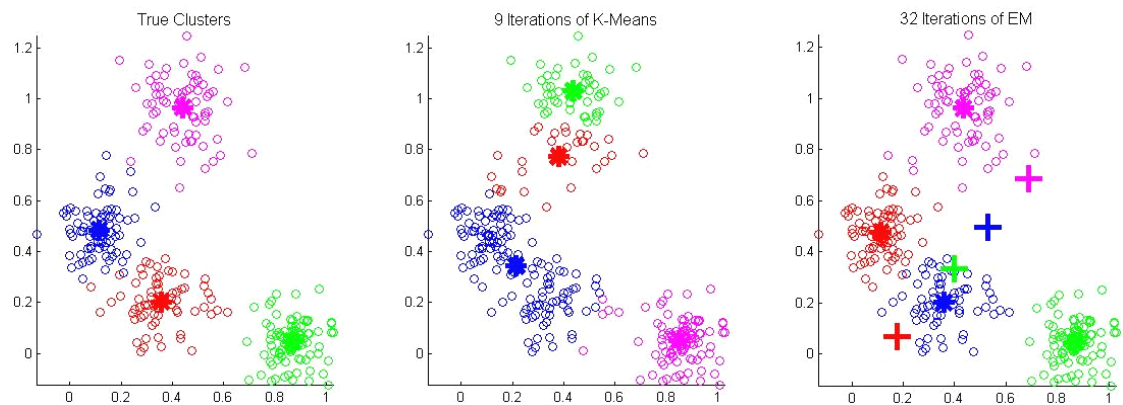

Figure 2: Case of four clusters in 2-dimension space and the *'s represent the mean for every cluster, and the +'s represent the initial mean for EM. In this case, the K-means performs poorly, while EM performs well.

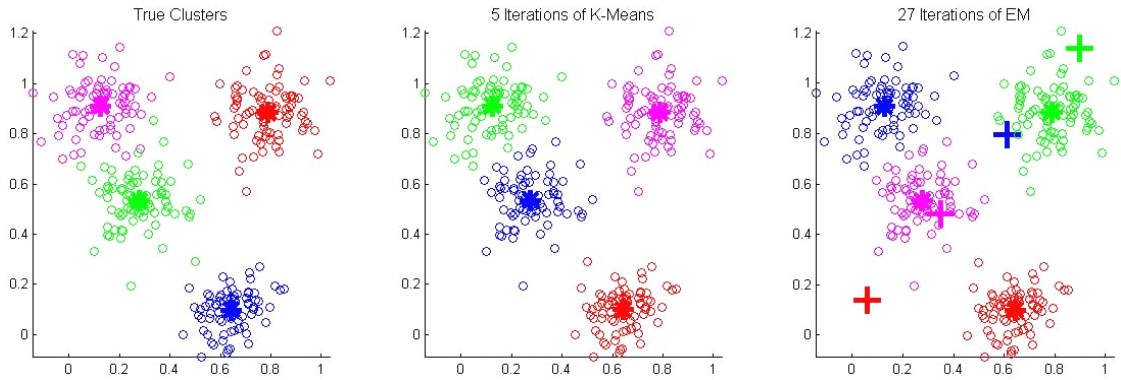

Figure 3: Case of four clusters in 2-dimension space and the *'s represent the mean for every cluster, and the +'s represent the initial mean for EM. In this case, both K-means and EM performs well.

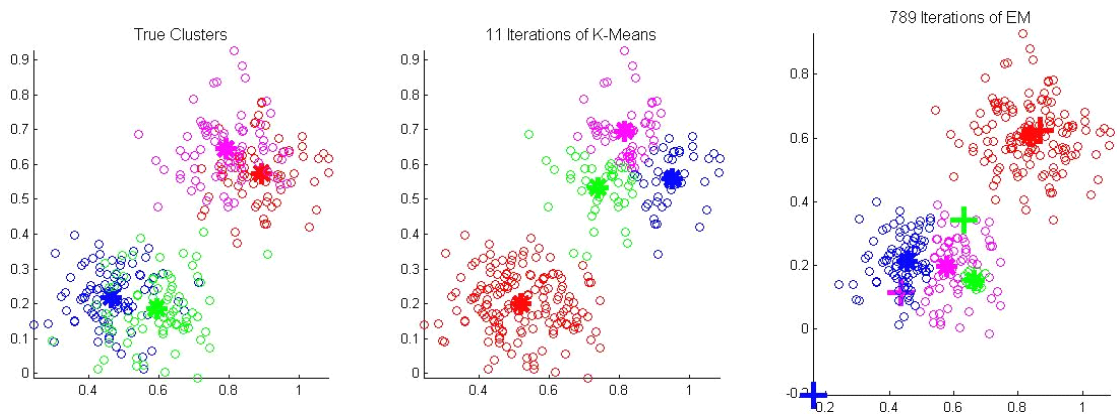

Figure 4: Case of three clusters in 2-dimension space and the *'s represent the mean for every cluster, and the +'s represent the initial mean for EM. In this case, none of EM and K-means perform well.

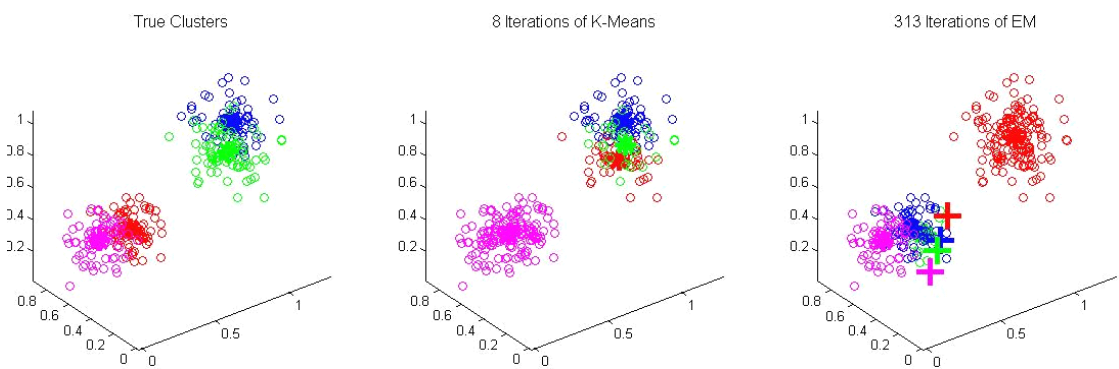

Figure 5: Case of four clusters in 3-dimension and the *'s represent the mean for every cluster, and the +'s represent the initial mean for EM. In this case, none of EM and K-means perform well.

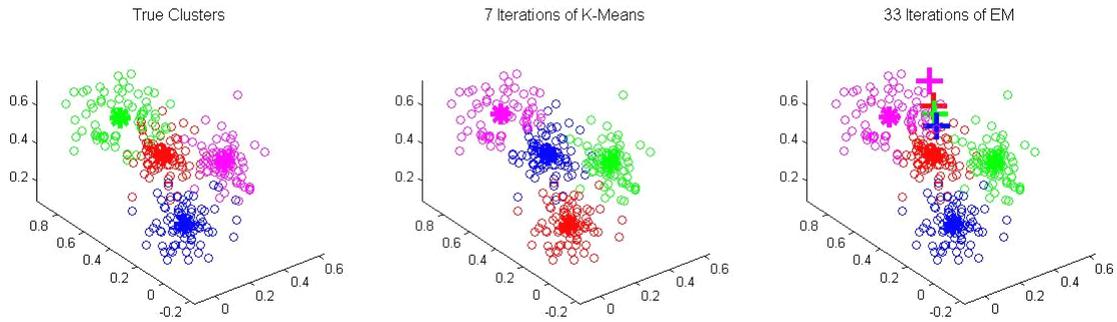

Figure 6: Case of four clusters in 3-dimension and the *'s represent the mean for every cluster, and the +'s represent the initial mean for EM. In this case, both EM and K-means perform well.

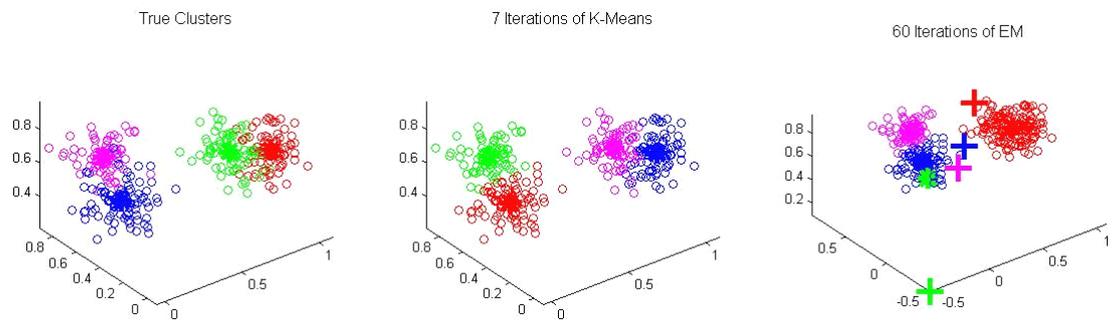

Figure 7: Case of four clusters in 3-dimension and the *'s represent the mean for every cluster, and the +'s represent the initial mean for EM. In this case, the K-means performs well, while EM performs poorly.

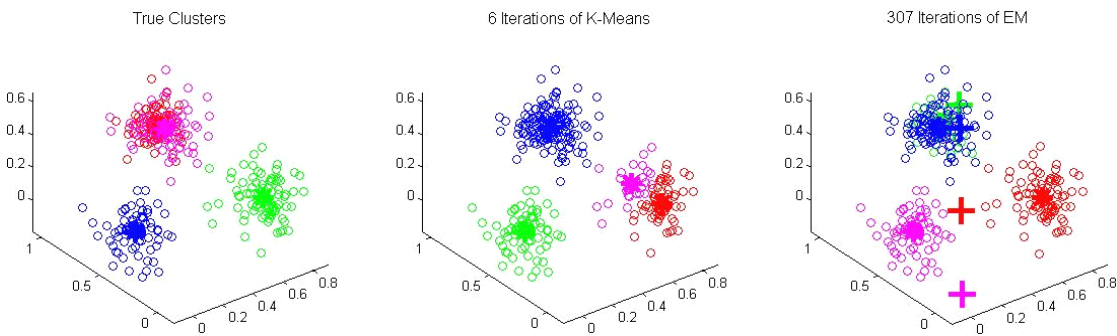

Figure 8: Case of four clusters in 3-dimension and the *'s represent the mean for every cluster, and the +'s represent the initial mean for EM. In this case, the EM performs well, while K-means performs poorly.